\documentclass{bmvc2k}
\usepackage{times}
\usepackage{epsfig}
\usepackage{graphicx}
\usepackage{amsmath}
\usepackage{amssymb}
\usepackage{multirow}

\usepackage{subcaption}
\usepackage{titlesec}
\titlespacing*{\section}{0pt}{0.1\baselineskip}{\baselineskip}
\title{E2ETag: An End-to-End Trainable Method for Generating and Detecting Fiducial Markers}

\addauthor{J. Brennan Peace}{jpeace@huskers.unl.edu}{1}
\addauthor{Eric Psota}{epsota@unl.edu}{2}
\addauthor{Yanfeng Liu}{yanfeng.liu@huskers.unl.edu}{3}
\addauthor{Lance C. P\'{e}rez}{lperez@unl.edu}{4}
\addinstitution{
 University of Nebraska-Lincoln\\
 Lincoln, Nebraska 68588-0511\\
}
\runninghead{Peace et al}{E2E Trainable Method for Generating and Detecting Markers}
\begin{document}

\maketitle
%%%%%%%%% ABSTRACT
\begin{abstract}
Existing fiducial markers solutions are designed for efficient detection and decoding, however, their ability to stand out in natural environments is difficult to infer from relatively limited analysis. Furthermore, worsening performance in challenging image capture scenarios - such as poor exposure, motion blur, and off-axis viewing - sheds light on their limitations. E2ETag introduces an end-to-end trainable method for designing fiducial markers and a complimentary detector. By introducing back-propagatable marker augmentation and superimposition into training, the method learns to generate markers that can be detected and classified in challenging real-world environments using a fully convolutional detector network. Results demonstrate that E2ETag outperforms existing methods in ideal conditions and performs much better in the presence of motion blur, contrast fluctuations, noise, and off-axis viewing angles.  Source code and trained models are available at \url{https://github.com/jbpeace/E2ETag}.
\end{abstract}
%\IEEEpeerreviewmaketitle
%%%%%%%%% BODY TEXT
\section{Introduction}
\vspace{-0.4cm}
Visual tracking aims to locate targets as they move through the field of view, while maintaining a consistent identification as targets disappear, reappear, and change their appearance \cite{bernardin2008evaluating}. The identity assigned to targets is, in general, arbitrary and their exact location is often represented via a bounding box \cite{redmon2017yolo9000} or a collection of key points \cite{cao2018openpose, Kristan2019a}. 

To obtain the precise location, identity, and pose of targets, one can use fiducial markers, which are man-made objects designed to to be placed in (augment) a scene. 
Along with algorithms to detect and classify them, they provide a plug-and-play tracking method that is scene-agnostic. 
Perhaps the most well-known fiducial markers are QR codes. 
When placed conveniently in front of a camera, QR codes can be detected and decoded efficiently \cite{liu2008recognition}. 
They are so ubiquitous that most modern cell-phone camera applications instantly recognize and decode them by default. QR codes are capable of encoding thousands of bits of information, but they are not designed to overcome difficult viewing conditions.

\begin{figure}
    \begin{center}
    \includegraphics[width=1.0\linewidth]{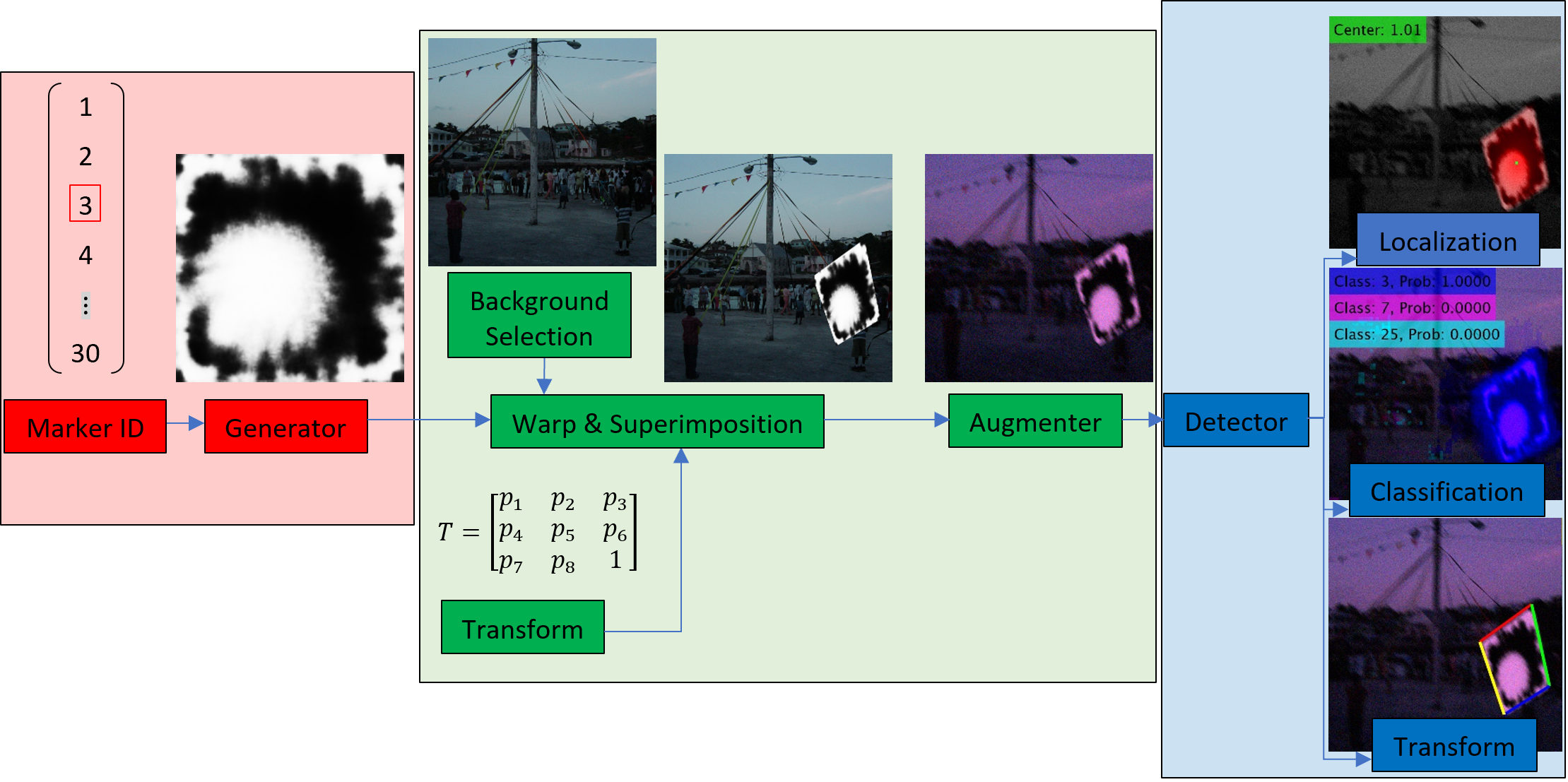}
    \end{center}
       \caption{Simplified model flowchart of E2ETag used to generate and detect fiducial markers.}
       % [p1,p2,p3;p4,p5,p6;p7,p8,1] = ...
       % [0.5829, -1.1126, 164.000; 
       %  0.8242, -0.5934, 96.0000; 
       %  0.0005,  0.0003, 1.0000];
    \label{fig:flow}
    \vspace{-0.5cm}
\end{figure}

This work targets a sub-category of fiducial markers aimed at challenging image capture scenarios.
Within this category, several methods have been proposed by the research community \cite{bergamasco2011rune,wang2016apriltag,degol2017chromatag}. Nearly all of them use two-dimensional bit encoding and heuristically designed detectors. While computationally efficient and reliable against false detections, they are not explicitly designed to handle real-world challenges like motion blur and noise.

The fiducial marker method introduced here takes a machine-learnable approach to both marker generation and detection. It relies on three stages that are end-to-end trainable, as illustrated in Figure \ref{fig:flow}. The first stage generates the fiducial marker from a one-hot vector using a transposed convolution. The second stage randomly augments the marker and superimposes it into a real image. Finally, the third stage uses a fully-convolutional network to detect the location, identity, and pose of the marker. The detector and generator learn to adapt to  severe augmentations and differentiate the marker from objects in real-world environments.

%The primary contribution is a method for generating robust fiducial markers using simulated superimposition in an end-to-end trainable network. To demonstrate that this approach works in practice, it is compared to existing state-of-the-art approaches. Public datasets and benchmarks do not exist for comparing new fiducial marker methods to existing markers. To do so would require either simulation or static cameras and static environments where markers are physically swapped in and out of the scene. Therefore, a new dataset is generated that includes indoor and outdoor scenes with varied scaling, rotation, and off-axis viewing. The proposed method outperforms existing fiducial marker systems in real-world environments and performs particularly well in the presence of image degradation. %

The contributions of this work include:
\begin{itemize}
    \item An end-to-end trainable framework for generating and detecting fiducial markers.
    \item Randomized, backpropagatable superimposition for simulated image capture.
    \item Analysis on the effectiveness of synthetic training for real-world applications.
    \item A methodology for evaluating the robustness of fiducial markers.
\end{itemize}

\vspace{-0.2cm}
\section{Related Work} 
\begin{figure}
\begin{center}
\begin{tabular}{cccc}
 \includegraphics[width=0.12\linewidth]{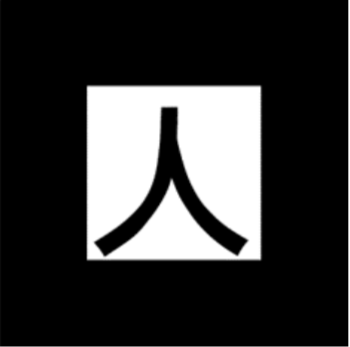} &   
 \includegraphics[width=0.12\linewidth]{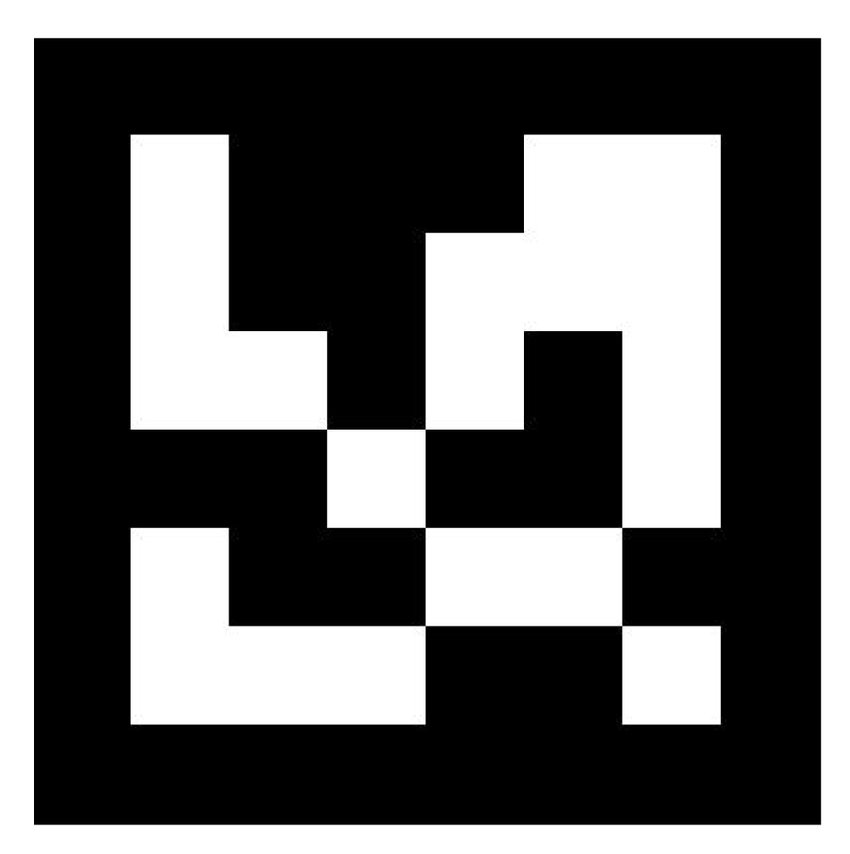} &
 \includegraphics[width=0.12\linewidth]{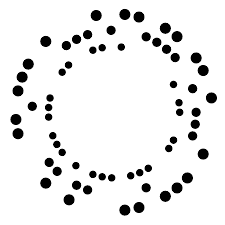} &
 \includegraphics[width=0.12\linewidth]{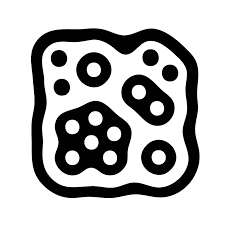} \vspace{-0.1cm} \\
 (a) ARToolkit & (b) ARTag & (c) RUNEtag & (d) reacTIVision
 \end{tabular}
 \vspace{-0.1cm}
 \begin{tabular}{ccccc}
 \includegraphics[width=0.12\linewidth]{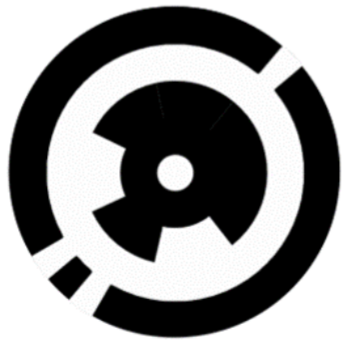} &
 \includegraphics[width=0.12\linewidth]{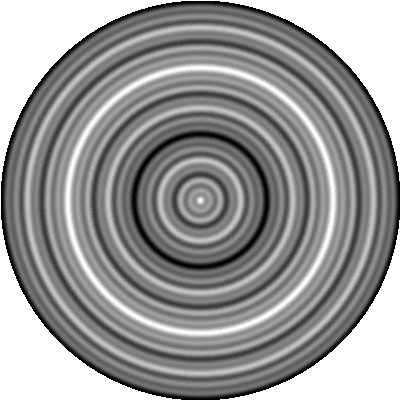}  & \includegraphics[width=0.12\linewidth]{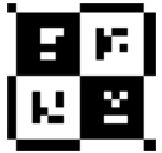} &
 \includegraphics[width=0.12\linewidth]{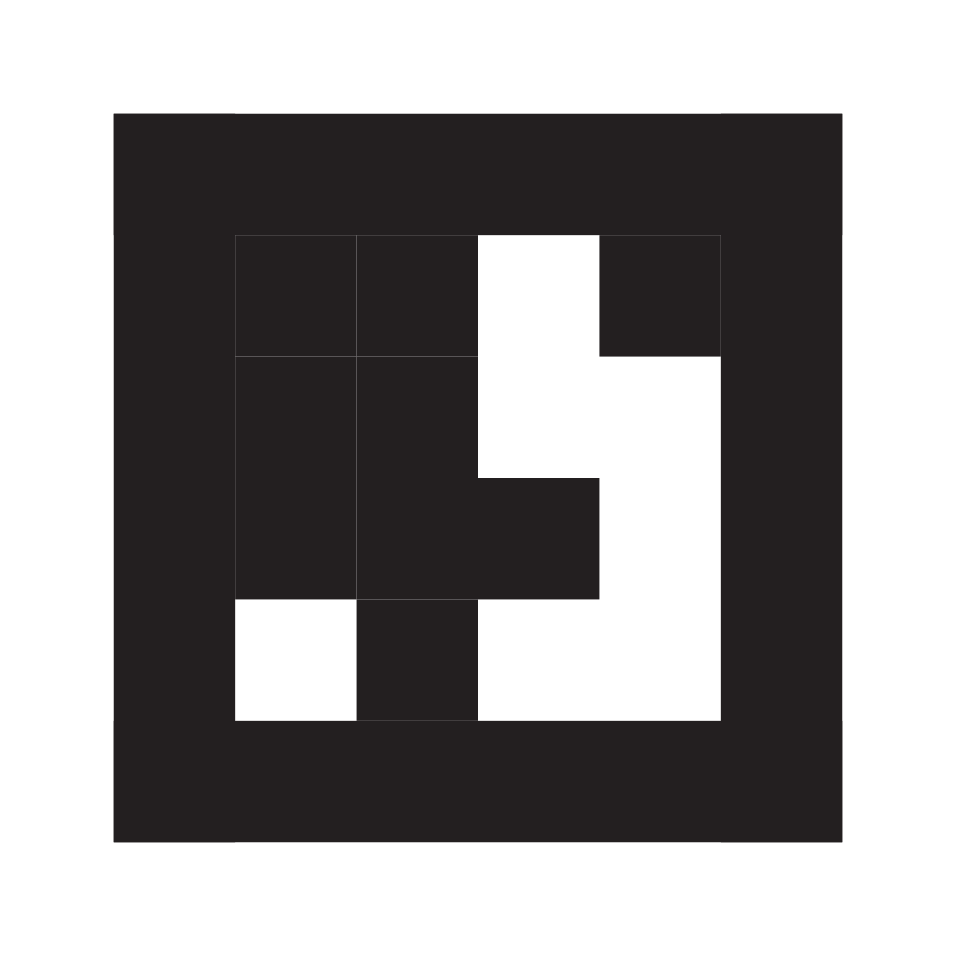} &   
 \includegraphics[width=0.12\linewidth]{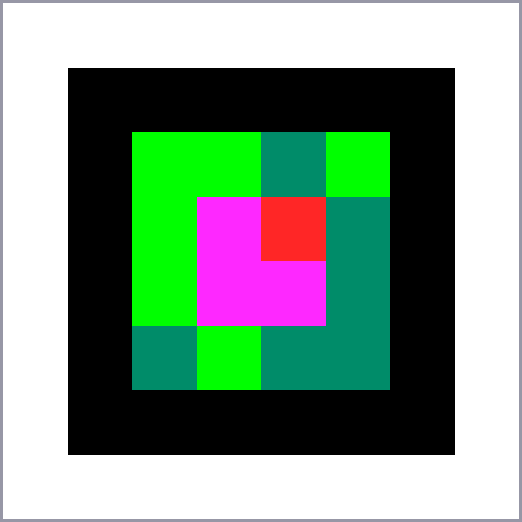} \vspace{-0.1cm}\\
 (e) BullsEye & (f) FourierTag & (h) CALTag & (i) AprilTag & (j) ChromaTag
\end{tabular}
\vspace{0.4cm}
\caption{Examples of existing fiducial marker designs.}
\label{fig:ex}
\end{center}
\vspace{-0.8cm}
\end{figure}

\vspace{-0.4cm}
Fiducial markers are designed to stand out from the environment and achieve a high detection rate while being, at the same time, distinguishable from one another for multi-tag detection. Traditionally, designs use a pre-determined library of decodable square patterns. 
ARToolKit \cite{kato1999marker}, one of the earliest fiducial markers, features an arbitrary pattern enclosed by a black border (Figure \ref{fig:ex}(a)). Performance is limited by the number of patterns and the camera resolution. The arbitrary nature of its content also makes inter-tag classification difficult to guarantee. ARTag \cite{fiala2005artag} proposed to fix this by using binary block patterns and introducing error-correction coding into its design (Figure \ref{fig:ex}(b)). 

RuneTag \cite{bergamasco2011rune}  (Figure \ref{fig:ex}(c)) exploits the projective properties of circular dot patterns and error-correction coding. The dots form one or more concentric circles, and the fact that both rings and dots appear elliptical under projective transformation makes decoding straight-forward. This design is robust under partial occlusion, blur, and noise. 
The authors of reacTIVision \cite{bencina2005improved} proposed topology-based irregular shapes for fast detection (Figure \ref{fig:ex}(d)). It supports a large number of markers, with a size that changes based on the number of encoded features. The design was originally proposed for table-based musical instruments \cite{kaltenbrunner2007reactivision} but can serve as a general-purpose marker. 
BullsEye is another topological pattern targeting the same applications as reacTIVision \cite{klokmose2014bullseye} (Figure \ref{fig:ex}(e)). 
It improves upon the precision  of reacTIVision and introduces GPU-enabled detection, however, both of these methods target two-dimensional location and orientation and require multiple markers for pose estimation.

FourierTag \cite{xu2011fourier} encodes information in the amplitude of a marker's Fourier transform (Figure \ref{fig:ex}(f)). It is designed to gradually degrade the quality of encoded information as distance increases and/or viewing angles worsen, instead of being unrecognizable abruptly. The high-order bits are encoded with low frequencies and low-order bits with high frequencies. As a result, the encoded bits have variable length depending on the viewing distance. 

CALTag \cite{atcheson2010caltag} (Figure \ref{fig:ex}(h)) proposes a high-density marker design as an alternative to checkerboard patterns and uniquely identifiable markers, with the application of camera calibration in mind instead of augmented reality. It offers an automatic processing procedure without parameter fine-tuning, which benefits multi-camera applications. 

AprilTag \cite{olson2011apriltag, wang2016apriltag} was designed to improve upon ARTag (Figure \ref{fig:ex}(i)). It proposes a grid of black and white blocks that serve as a binary payload with guaranteed minimum Hamming distance between markers undergoing 0, 90, 180, and 270 degree rotations. It was originally designed to handle partial-occlusion recovery, but the authors concluded that occluded markers were rarely useful and they instead targeted detection and decoding speed. 

ChromaTag \cite{degol2017chromatag} features adjacent red and green blocks surrounded by black and white rings (Figure \ref{fig:ex}(j)). The red and green pattern is rare in natural scenes and reduces initial false detections. The black and white rings provide high contrast for localization. In the CIELAB color space, each color in  the design has a consistent value in the A channel and a different value in the B channel, making the design easy to detect and decode.

Existing designs use hand-crafted patterns and detection algorithms. It is unclear if  detection is optimized for the marker or vice versa. To the best of our knowledge, the method introduced in this paper is the first end-to-end trainable fiducial marker solution. The marker designs are jointly optimized with their detector, allowing for designs that learn to stand out from the environment while simultaneously learning to look different than each other. 
\vspace{-0.3cm}
\section{Method}
\vspace{-0.4cm}
The model used during training is composed of a three-stage generator/augmenter/detector network (Figure~\ref{fig:flow}). The first stage generates markers through a transposed convolution. The second stage warps them onto a sample image and applies augmentations that simulate real-world image capture.  The final stage estimates the marker's location, class, and pose.
%-------------------------------------------------------------------------
\vspace{-0.2cm}
\subsection{Generator}
\vspace{-0.2cm}
The generator is defined by a transposed convolution where the input is a one-hot vector indicating the desired class.  The transposed convolution consists of a single kernel of size $S\times S\times C$, where $S = 128$ and $C = 30$. When the $1\times 1\times C$ one-hot vector is convolved with the kernel, the $S\times S \times 1$ output is a single kernel layer representing an E2ETag image.
%-------------------------------------------------------------------------
\vspace{-0.3cm}
\subsection{Spatial Warp and Superimposition}
\vspace{-0.2cm}
The E2ETag image is warped and superimposed into a real image. Background images are randomly sampled from the COCO \cite{lin2014microsoft} and Imagenet \cite{deng2009imagenet} data sets and resized to $640 \times 640$.  The spatial warping transform is constructed by randomly generating $x$ and $y$ translations $\{t_x, t_y\}$ ranging $(0,640)$, rotation $r$ ranging $(0,2\pi]$, scaling $\{s_x, s_y\}$ ranging $(8/128,320/128)$, shear $\{h_x, h_y\}$ ranging $(-3\pi/12,3\pi/12)$, and projective warping $\{w_x, w_y\}$ ranging from $(-0.0015,0.0015)$.  Additionally, a shift of $-S/2$ is applied to both dimensions to zero center the marker at the origin prior to warping. The resulting projective matrix is 
\begin{equation}
\mbox{\footnotesize $
T = 
\begin{bmatrix} 1 \hspace{-0.2cm}& 0 \hspace{-0.2cm}& t_x \\ 0 \hspace{-0.2cm}& 1\hspace{-0.2cm} & t_y \\ 0 \hspace{-0.2cm}& 0 \hspace{-0.2cm}& 1 \end{bmatrix} \hspace{-0.12cm}\times\hspace{-0.12cm} \begin{bmatrix} 1 \hspace{-0.2cm}& 0 \hspace{-0.2cm}& 0 \\ 0 \hspace{-0.2cm}& 1 \hspace{-0.2cm}& 0 \\ w_x \hspace{-0.2cm}& w_y \hspace{-0.2cm}& 1 \end{bmatrix} \hspace{-0.12cm}\times\hspace{-0.12cm}
\begin{bmatrix} \cos(r) \hspace{-0.2cm}& -\sin(r) \hspace{-0.2cm}& 0 \\ \sin(r) \hspace{-0.2cm}& \cos(r) \hspace{-0.2cm}& 0 \\ 0 \hspace{-0.2cm}& 0 \hspace{-0.2cm}& 1 \end{bmatrix}\hspace{-0.12cm}\times\hspace{-0.12cm} 
\begin{bmatrix} 1 \hspace{-0.2cm}& h_y \hspace{-0.2cm}& 0 \\ h_x \hspace{-0.2cm}& 1 \hspace{-0.2cm}& 0 \\ 0 \hspace{-0.2cm}& 0 \hspace{-0.2cm}& 1 \end{bmatrix}\hspace{-0.12cm}\times\hspace{-0.12cm}
\begin{bmatrix} s_x \hspace{-0.2cm}& 0 \hspace{-0.2cm}& 0 \\ 0 \hspace{-0.2cm}& s_y \hspace{-0.2cm}& 0 \\ 0 \hspace{-0.2cm}& 0 \hspace{-0.2cm}& 1 \end{bmatrix}\hspace{-0.12cm}\times\hspace{-0.12cm}
\begin{bmatrix} 1 \hspace{-0.2cm}& 0 \hspace{-0.2cm}& -S/2 \\ 0 \hspace{-0.2cm}& 1 \hspace{-0.2cm}& -S/2 \\ 0 \hspace{-0.2cm}& 0 \hspace{-0.2cm}& 1 \end{bmatrix}
= \begin{bmatrix} p_1 \hspace{-0.2cm}& p_2 \hspace{-0.2cm}& p_3 \\ p_4 \hspace{-0.2cm}& p_5 \hspace{-0.2cm}& p_6 \\ p_7 \hspace{-0.2cm}& p_8 \hspace{-0.2cm}& 1 \end{bmatrix}.
$}
\end{equation}
The parameters corresponding to rotation, scaling, and shearing $(p_1,p_2,p_4,p_5,p_7,p_8)$ are used by the loss function for training, as discussed in section ~\ref{training}.
%-------------------------------------------------------------------------
\vspace{-0.3cm}
\subsection{Local and Pixel-Level Augmentations}
\begin{figure*}
    \begin{center}
    \includegraphics[width=0.95\linewidth]{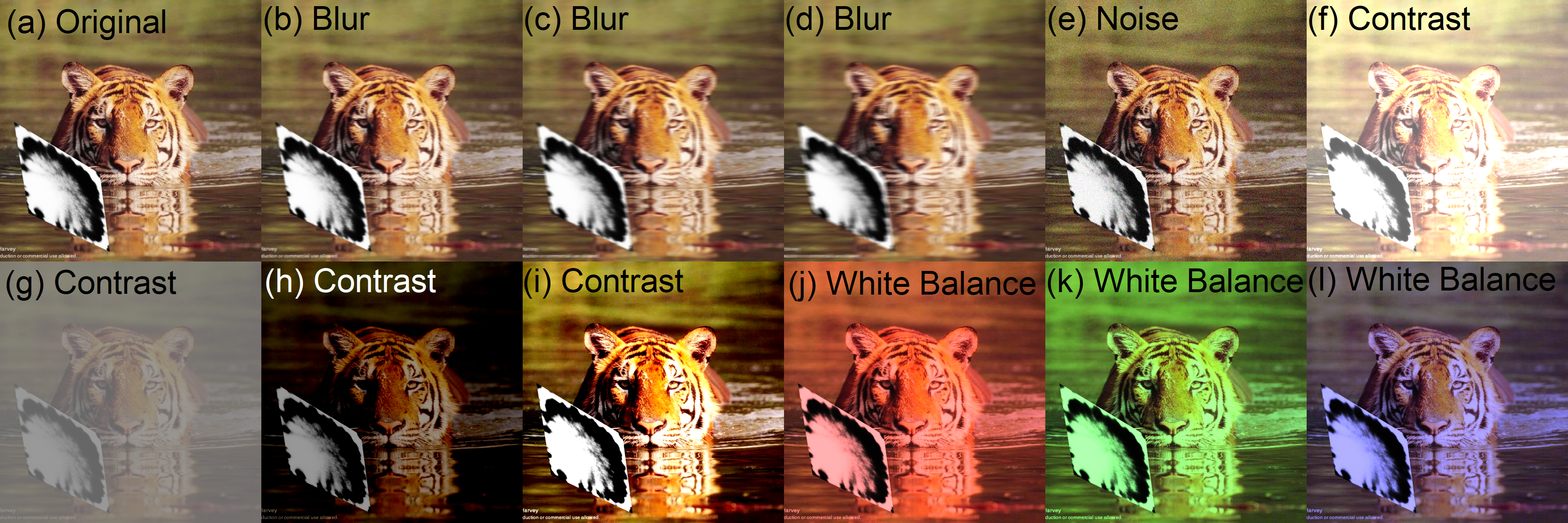}
    \end{center}
    \caption{Superimposed marker without augmentations (Figure3a). Motion Blur at random angles with varying kernel length 5, 10, and 15 (Figure3 b,c,d).  Additive noise ranging (-0.15,0.15) (Figure3e). Contrast with $W=1.4$ and $B=0.4$ (Figure3f), $W=0.6$ and $B=0.4$ (Figure3g), $W=0.6$ and $B=-0.4$, and $W=1.4$ and $B=-0.4$ (Figure3i).  White balance $(1.3,0.7,0.7)$ (Figure3j),  $(0.7,1.3,0.7)$ (Figure3k) and $(0.7,0.7,1.3)$ (Figure3l).}
    \label{fig:aug_all}
    \vspace{-0.5cm}
\end{figure*}

\vspace{-0.2cm}
Four separate augmentations are applied sequentially to images with superimposed markers: motion blur, white-balance, contrast, and additive noise, as illustrated in Figure~\ref{fig:aug_all}.
Motion blur is simulated first by convolving a blur kernel with variable angle and length.  This kernel simulates linear camera motion along a direction $d$ uniformly ranging $(0,2\pi]$ with pixel length $l$ uniformly ranging $(0,10)$ pixels.  Input Image $I$ is convolved with kernel $\phi(d,l)$ to produce the output image $I_{MB} =  I \ast \phi(d,l)$.

White balance simulates lighting conditions with varied temperatures.  Random values for each channel, uniformly ranging $(0.7,1.3)$, scale each channel of an RGB input image. The resulting image is $I_{WB} = I_{MB}\odot C$, where $C$ is an image of the same size as $I_{MB}$ where each channel is given a single scale value and $\odot$ is an element-wise product.

A contrast augmentation is then used to simulate variations in exposure, post processing, and light pollution. Contrast considers two random variables: the maximum white $W$  uniformly distributed  $(0.6, 1.4)$ and minimum black $B$  uniformly ranging $(-0.4, 0.4)$. Contrast is adjusted using $I_{C} = I_{WB}\times (W-B) + B$.

Finally, a noise image $N$ with pixel values uniform in the range $(-n/2,n/2)$ is added to the input image. Image $I_C$ is augmented with noise using $I_N = I_{C} + N$. Finally, pixel values in $I_N$ are clipped between 0 and 1.  
%\vspace{-0.2cm}
%\subsection{Dual Contrast and Superimposed Images}
%\vspace{-0.2cm}
%Prior to superimposing E2ETag onto the background image, the marker is augmented with contrast.  Dual Contrast augmentation is applied to ensure the detector considers spatially-varying contrast.  This includes lighting reflections, shadows, and multiple lighting sources simulating a multi-modal intensity distributed throughout the image.  This was discovered heuristically after noticing that the detector performed better than original capture when the image was further augmented with contrast.
%After superimposing E2ETag and applying all other augmentations, the image is superimposed with random COCO images.  The number of superimposed images is random ranging from (0,10).  The transform of these images are either the same as E2ETag with different random translation values or the transform is randomly sampled.  Introducing additional superimposed images that were not E2ETag was intended to encourage the detector to ignore other objects that appear superimposed and to instead force the detector to recognize features pertaining only to E2ETag.
%-------------------------------------------------------------------------
\vspace{-0.45cm}
\subsection{Detector}
\vspace{-0.3cm}
The final stage of the model, used for both training and testing, is the detector. The detector localizes the marker within the image, classifies it, and estimates the transformation parameters.  The pre-trained, fully-convolutional network chosen in this work is the DeepLabV3+ architecture \cite{chen2018encoder} with a ResNet18 core \cite{he2016deep}. This network was chosen due to its speed, classification performance, ability to handle multi-scale detection, and high resolution feature space.
Instead of up-sampling back to the input resolution, the output before the first transposed convolution is used and the resulting network down-samples the input image scale from $640\times640$ to $80\times 80$.
%The output of the augmented markers is normalized using values for a trained ResNet18 with set mean $\mu$ and standard deviation $\sigma$ vectors.  The mean $\mu = [0.485, 0.456, 0.406]$ and the standard deviation $\sigma = [0.229, 0.224, 0.225]$ are applied to the input image into the detector by normalizing to achieve the output image $I_{Norm}$:
%\begin{equation}
%I_{Norm} = (I_N-\mu)/\sigma
%\end{equation}
A $1\times1\times256\times256$ convolution/batchnorm/ReLU block and $1\times1\times256\times37$ convolution layer are added to provide an output with the number of channels used to encode targets. 
Those channel encodings are described in the following.
\begin{figure}
    \begin{center}
    \includegraphics[width=0.75\linewidth]{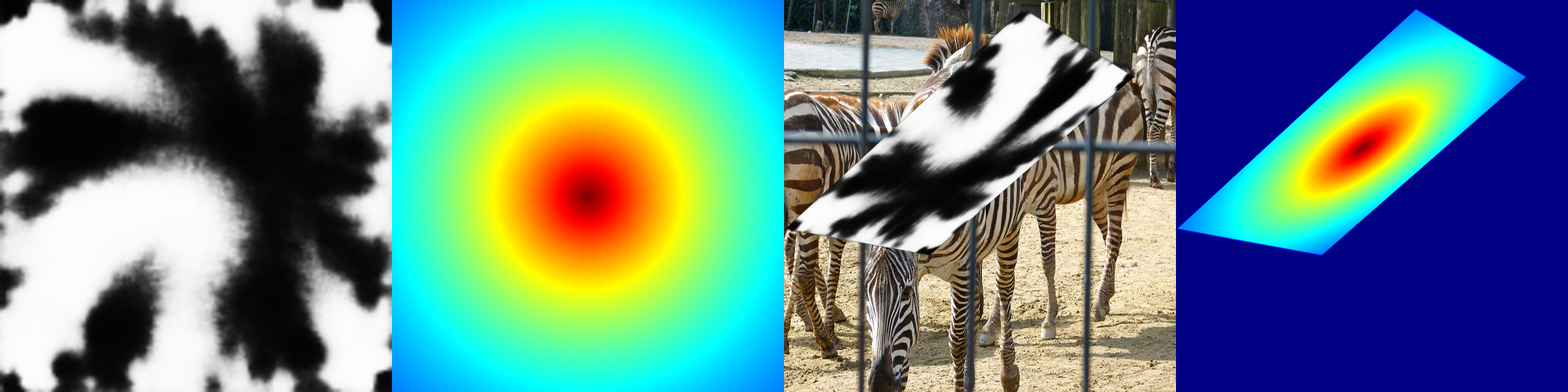}
    \end{center}
       \caption{Sample E2ETag and decaying exponential used to encode marker location. Both marker and location encoding are superimposed the same way into the image. In the location encoding, dark red indicates a value of 1.0 and dark blue indicates a value of 0.0.}
    \label{fig:dec_exp}
    \vspace{-0.6cm}
\end{figure}
%-------------------------------------------------------------------------
\vspace{-0.45cm}
\subsubsection{Detector Channel Encoding}
\vspace{-0.25cm}

The first channel, of size $80\times 80\times 1$, is used to detect and localize the marker. 
The marker's location in the image is encoded via a decaying exponential setting the value at each pixel to $e^{\text{-}(r/64)}$, where $r$ is the distance of each pixel from the center. 
The decaying exponential image has the same size as the marker with dimensions $128\times 128$.
Figure \ref{fig:dec_exp} illustrates a marker image and its corresponding mapping in the first channel before and after warping and superimposing. 
%The centroid of the marker is determined by averaging the location of the corners in the original image and then dividing by the downsampling factor of the detector network (e.g., 32 for ResNet18). The distance between this real-valued coordinate and the centers of all feature space coordinate centers is calculated. The first channel of the output is then encoded with the value of one minus this distance lower bounded to the value of zero.  In practice, this means that the maximum value in the first channel is one, and that a maximum of four feature space coordinates have value greater than zero for each marker present in the image. 
To detect tag locations, the method begins by finding all regional maxima in 3$\times$3 regions in the $80\times 80\times 1$ output. 
If the value of the regional maxima exceeds 0.5, it is considered a tag center and its sub-pixel peak is estimated using quadratic interpolation.

The next 30 channels, with feature space size $80\times 80\times 30$, encode marker identities using softmax pixel-wise classification. Classification is trained equally for all pixels occupied by the transformed marker. All other grid locations are allowed to choose identities without affecting the loss, so that uncertainty in detection does not affect classification.

The last six channels of the output, with feature space size $80\times 80\times 6$, contain the projective parameters $(p_1,p_2,p_4,p_5,p_7,p_8)$ at each pixel location occupied by the marker.
\vspace{-0.4cm}
\subsection{Training Details}
\begin{figure}
    \begin{center}
    \includegraphics[width=1.0\linewidth]{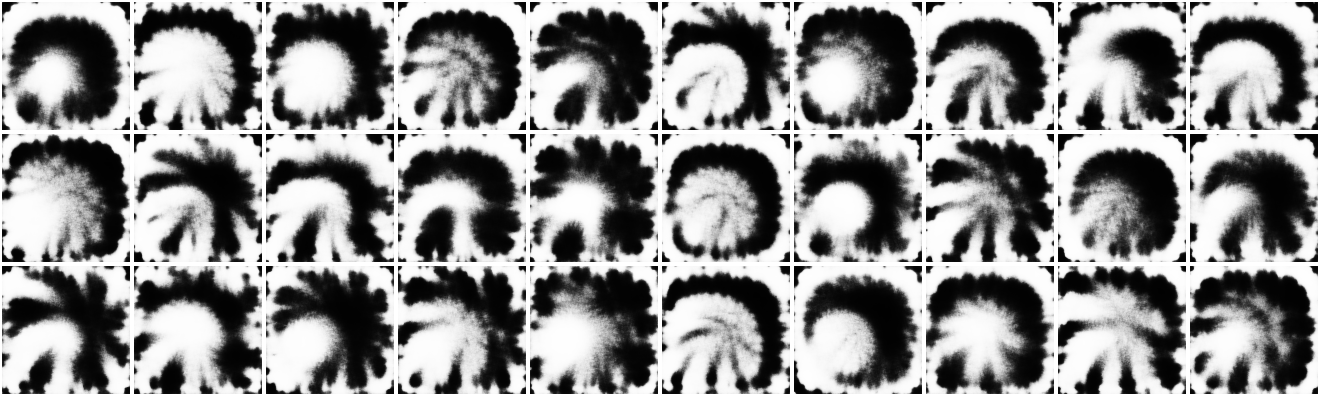}
    \end{center}
       \caption{E2ETag markers generated with 30 classes.}
    \label{fig:marker}
    \vspace{-0.5cm}
\end{figure}

\vspace{-0.2cm}
\label{training}
The network is end-to-end trainable through the detector, augmentation, and warp layers. While improving the detector, training also encourages marker designs that are easy to detect and classify. 
While the detector was mostly pre-trained, the transposed convolution weights used to generate markers were randomly initialized with Gaussian samples that have mean 0.0 and variance 1.0. The bias was fixed to zero. Outputs of the transposed convolution are passed through a preliminary contrast augmentation layer then to a sigmoid layer to limit values between 0 and 1 prior to superimposing into images.  Because  early layers typically train much more slowly than later layers, the learning rate of the transposed convolution was increased by a factor of 1000. 
The Adam optimizer was used with a learning rate of $2\times 10^{-5}$, a batch size of 8, and an L2 regularization of $10^{-4}$.
%The transposed convolutions also had a problem of exploding gradients, since we passed them into a sigmoid layer before being warped onto an image, it was possible for their values to get stuck in the asymptotic regions of the sigmoid function, being largely negative or largely positive.  To combat this, we added an L2 regularizer with value $1e^{-4}$ on the weights of the transposed convolution.  This prevented the outputs of the transposed convolutions from becoming largely positive or negative.  This allows gradients within the network to have a larger impact on marker design. 
The complete set of markers used in evaluation are depicted in Figure~\ref{fig:marker}.

\vspace{-0.2cm}
\subsection{Backward Propagation Through Warping and Augmentation}
\vspace{-0.2cm}
When superimposing markers, transformations $T$ were accepted only if each marker corner remained inside the image bounds. 
For back-propagation, the inverse transformation $T^{\text{-}1}$ was used to map gradients back to their origins in the marker image. 
The inverse transform also ensures that gradients are only applied to the marker; background regions are ignored.
Both forward and backward transformations use bilinear interpolation.
%-------------------------------------------------------------------------
%\vspace{-0.2cm}
%\subsection{Backward Propagation: Augmentations}
%\vspace{-0.2cm}
%The two backpropagation functions discussed involve the motion blur augmentation and the warp function.  To perform the backward operation to update the weights of the transposed convolutions these operations need to retain the gradients passed from the back-end detector. 

Gradients passed through the motion blur operation are back-propagated by reusing the motion blur kernel on the gradients.  
Gradients for white balance and contrast adjustments are scaled according to their multiplicative factors. 
%Finally, additive noise does not affect the gradients.
%This is possible since the blur kernel is essentially the weighted sum of the kernel applied to the image and is differentiable.

%Gradients passing through the warp operation are passed using the same projective transform for warping the markers onto the background images.  The inverse of the transform parameters is used to warp the gradients within the warped marker location back to the raw marker locations.  The inverse transform only considers the regions of the warped marker due to the nature of the transform; background regions are ignored without any attention considered. The inverse warp operation is performed with bilinear interpolation in the forward and backward direction.
%-------------------------------------------------------------------------
\vspace{-0.2cm}
\subsection{Loss Function}
\vspace{-0.2cm}
%Loss is defined as an aggregate penalty of the classification, regression of localization, and transform parameters.  Detection/localization loss is penalized using MSE and classification loss is the categorical cross-entropy loss of target class versus the predicted class.  The projective transformation parameters are not penalized directly. Instead, the corners of the ground truth marker and the estimated marker are calculated from the estimated projective parameters.  Those two sets of corners are both normalized by the standard deviation of the ground truth corners to ensure scale invariant training.  Finally, the sum of squares between the normalized predicted and target corners is used as the loss function for projective transformation prediction.%
%%%LOSS FUNCTION%%%
Loss is the aggregate penalty of errors in classification, localization, and estimation of transform parameters.  Detection/localization loss, $L_\text{loc}$, is penalized as the mean-square error, given by
\begin{equation}
    L_{\text{loc}} =  \sum_{i=1}^{\text{Rows}} \sum_{j=1}^{\text{Cols}} {(y_{\text{loc}}(i,j) - \hat{y}_{\text{loc}}(i,j))}^2 / \text{NumPix},
\end{equation}
where $y_{\text{loc}}$ is the image target localizations, $\hat{y}_{\text{loc}}$ is the image with predicted localizations, and $\text{NumPix}$ is the number of pixels within the warped marker region.
\par
Classification loss is the categorical cross-entropy loss of target class versus the predicted class, given by
\begin{equation}
    L_{\text{class}} =  -\sum_{c=1}^{C} \sum_{i=1}^{\text{Rows}} \sum_{j=1}^{\text{Cols}} y_{\text{class}}(i,j,c)\odot \log(\max(\hat{y}_{\text{class}}(i,j,c),\epsilon)) / (C\cdot \text{NumPix}),
\end{equation}
where $C$ is the number of classes, $y_{\text{class}}$ is the image target classifications, $\hat{y}_{\text{class}}$ is the image with predicted classifications, and $\epsilon = 10^{\text{-}9}$ is a constant used to prevent division by zero.

The projective transformation parameters are not penalized directly; instead, the corners of the target marker and the predicted marker are calculated from the estimated projective parameters.  Parameters $p_3$ and $p_6$ are set to zero to isolate transformation from localization.  %The corners projected with the transformation are centered around the origin to encourage training to put an equal loss on all transform parameters.  For example, centering the marker before transforming prevents the loss from putting a heavy emphasis on rotation.  Without centering, the origin would experience no loss, while other corners experience varying degrees of loss.  The inconsistent corner loss empirically leads to divergent transformations that make training difficult.
The predicted corners $\hat{\mathbf{c}}$ and target corners $\mathbf{c}$ are both normalized by the standard deviation of the target corners $\sigma_c$ to produce $\mathbf{c}_N$ and $\hat{\mathbf{c}}_N$.  Normalizing the corners allows projective transforms to be penalized with scale invariance, preventing large markers from dominating training.  The $K$ strongest target localizations in $y_{\text{loc}}$ determined the grid locations to sample the transformations.  Here, the strongest $K=5$ detections were chosen to sample the true center and its four abutting neighbors.  Projective transformation loss is the mean-absolute error between normalized predicted and target corners, defined by
\begin{equation}
    L_{\text{proj}} = \sum_{k=1}^{K} \sum_{i=1}^{4} \sum_{j=1}^{2}  \left| \hat{\mathbf{c}}_{N}(i,j,k) - \mathbf{c}_{N}(i,j,k) \right| / (8\cdot k),
\end{equation}  
where $j$ is the index for an $(x,y)$ coordinate, $i$ is the index for the corner, and $k$ is the index for each localization maxima.
Finally, the total loss $L$ is an aggregate of all losses, given by 
\begin{equation}
    L =  a\cdot L_{\text{class}} +  b\cdot L_{\text{loc}} +  L_{\text{proj}},
\end{equation}
where $a=100$ and $b=50$ are scalar constants derived empirically to balance training.
%-------------------------------------------------------------------------
\vspace{-0.2cm}
\section{Results}
\vspace{-0.4cm}
E2ETag is compared to two state-of-the-art fiducial marker methods: ChromaTag and AprilTag.  
%Our marker is trained by a neural network with the possibility of improving results significantly.  
Two different metrics are used to evaluate performance: detection and classification. For detection accuracy, the intersection over union (IoU) derived from the corner locations is used and a true positive detection is defined by IoU greater than 50\% with the ground truth. 
Classification performance is simply evaluated as correct or incorrect class output.
%-------------------------------------------------------------------------
\vspace{-0.3cm}
\subsection{Data Collection}
\vspace{-0.2cm}
For each image used in the evaluation, a single ChromaTag, AprilTag, and E2ETag are placed in a scene with similar pose.  
Each of these methods was configured to support 30 different marker classes (AprilTag and ChromaTag use 16H5 encoding). 
%While it is assumed that the performance of detection and classification on a single sample is representative of the whole set. 
Images with all three markers in each frame are captured at $3024\times3024$ resolution with a 35mm equivalent focal length of 52mm.  They are captured in seven different environments, placed in varying lighting environments, and mounted on man-made structures as well as natural environments.  

The markers were attached to a cardboard square, 7.62$\times$7.62 cm, and mounted to different surfaces including trees, poles, a stone structure, a reflective glass window, and a chain-link fence. 
Seven different markers were used with identities 5, 6, 7, 9, 11, 19, and 27.
Twenty-five images from each environment were captured at five different angles ($-80^\circ$, $-40^\circ$, $0^\circ$, $40^\circ$, $80^\circ$) and five different distances (1, 2, 3, 4, and 5 meters).
The true corner locations were hand-annotated for each marker at the original resolution $3024\times3024$ and downsampled to $640\times640$.% and the corner coordinates adivided by 4.725 (the scale factor $3024/640$) and used as the ground truth locations.
%-------------------------------------------------------------------------
%\subsection{Detection}
%The proposed network infers the affine transform parameters, class, and centroid positions.  The grid distances are converted to centroid positions based off the regional maximum; thresholds determine the number detections.  For the regional maximum predictions, the affine transform parameters are used to calculate the predicted corners for each detection along with the highest confidence class prediction.  Figure~\ref{fig:show_our_det} conveys our class and center predictions along with the detected transform.  ChromaTag and AprilTag also process the same size image $640\times640$.  They each output their corner predictions, their predicted class, and the number of total detections made.
%-------------------------------------------------------------------------
\vspace{-0.3cm}
\subsection{Performance Comparison}
\begin{figure*}
    \begin{center}
    \begin{tabular}[c]{cc}
    \begin{subfigure}[c]{0.6\linewidth}
        \includegraphics[width=1.0\linewidth,trim=35 0 20 0,clip]{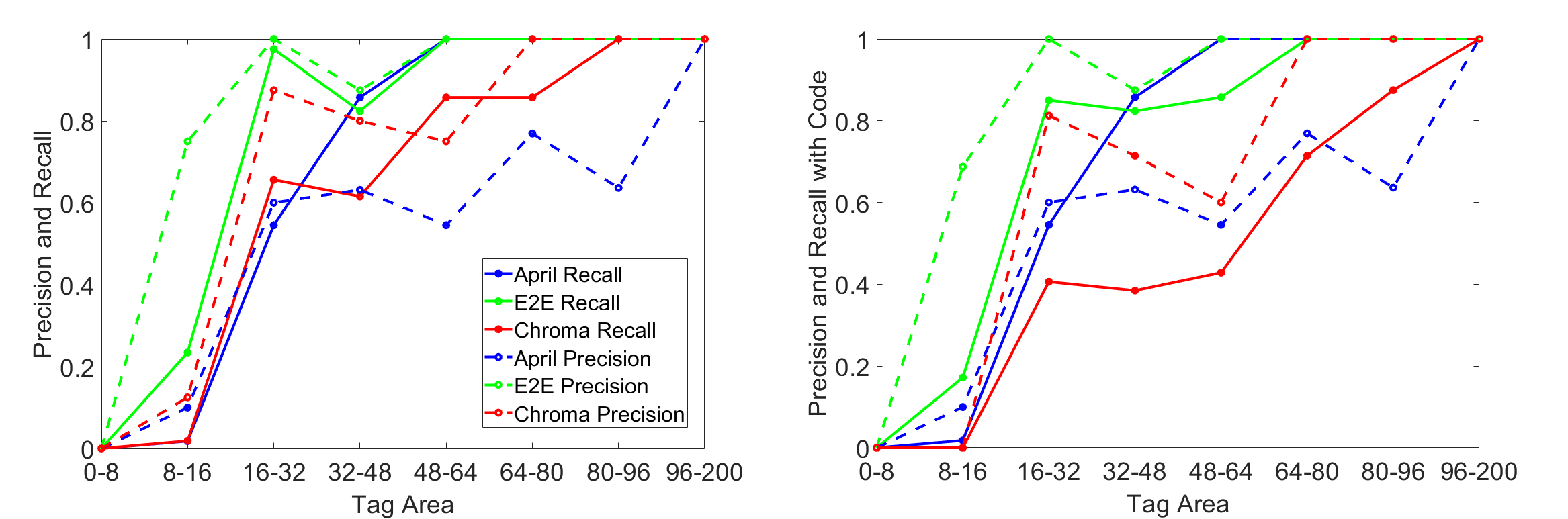}
        \caption{Precision and Recall vs Tag Area}
        \label{fig:recall_dist}
    \end{subfigure}
    &
    \multirow{2}{*}{\begin{subfigure}[c]{0.35\linewidth}
    \includegraphics[width=1.0\linewidth,trim=50 20 70 50,clip]{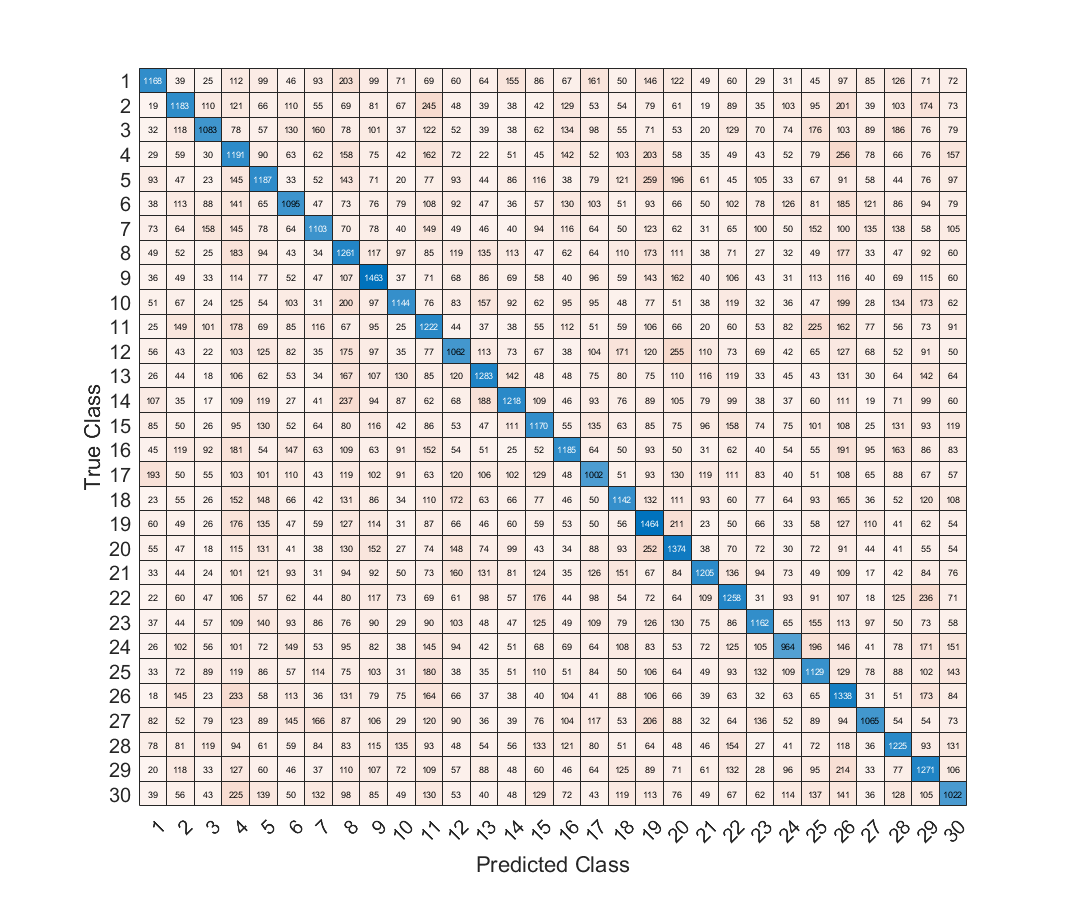}
        \caption{Confusion Matrix from Augmented Training Set}
        \label{fig:confmat}
    \end{subfigure}}
    \\
    \begin{subfigure}[c]{0.6\linewidth}
        \includegraphics[width=1.0\linewidth,trim=35 0 20 0,clip]{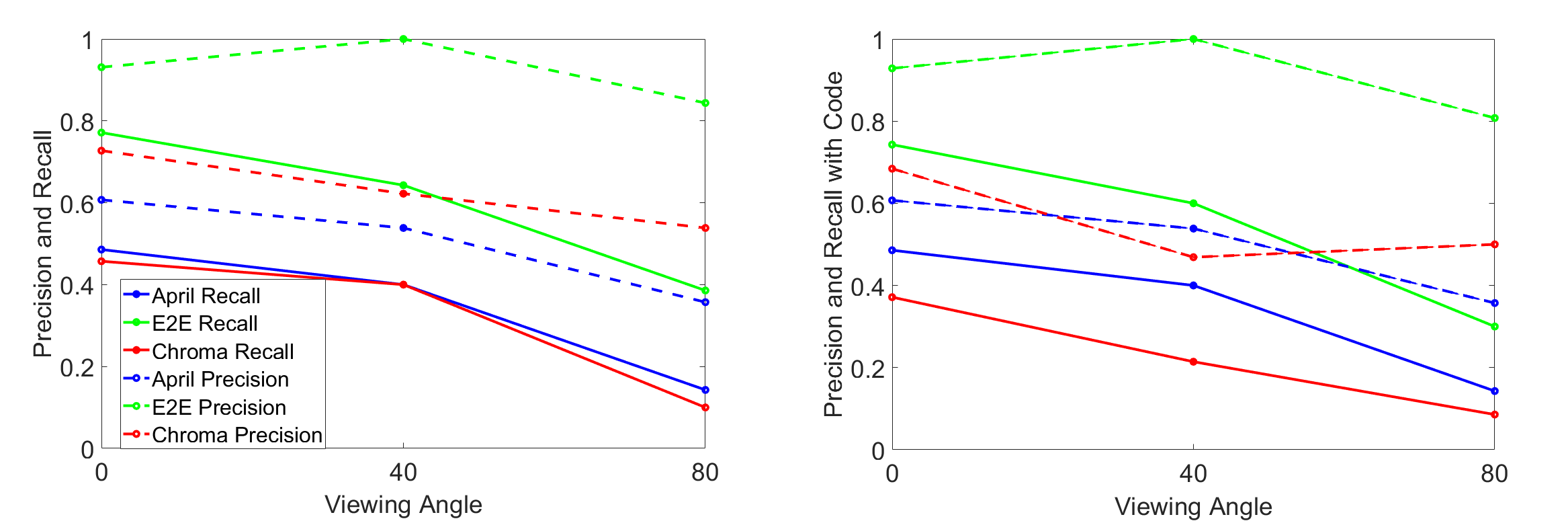}
        \caption{Precision and Recall vs Viewing Angle}
        \label{fig:prec_dist}
    \end{subfigure}
    \end{tabular}
    \end{center}
    \caption{Precision and Recall curves ((a) and (c)) across distance from the marker in meters and viewing angle of the marker in degrees.  For each graph \textcolor{blue}{AprilTag} is blue, \textcolor{red}{ChromaTag} is red, and \textcolor{green}{E2ETag} is green.  Precision and Recall are indicated by a dotted and solid line respectively. The confusion matrix for marker classification is also given in (b).}
    \label{fig:pr}
    \vspace{-0.5cm}
\end{figure*}

\vspace{-0.2cm}
The chosen metric for localization accuracy is the IOU of the quadrilaterals defined by the corners of the predicted and annotated corners.  Recall and precision are used to evaluate the performance of localization.
%Recall is defined as the ratio of True Positives over the sum of True Positives and False Negatives.  
%Precision is defined as the ratio of True Positives over the sum of True Positives and False Positives.  
True positives require $50\%$ IoU with the ground truth quadrilaterals.  False negatives are defined by an undetected ground truth marker.  Precision and recall versus distance from the marker and the viewing angle are shown in Figure~\ref{fig:pr}. 
%-------------------------------------------------------------------------
%\vspace{-0.2cm}
%\subsection{Classification Accuracy}
%\vspace{-0.2cm}

Classification accuracy is also evaluated using precision and recall.  In this case, only markers with the correct class prediction are counted toward the true positive rate.  Thus, a false positive has either less than $50\%$ IoU, incorrect class prediction, or both.
The results on the left side of Figure~\ref{fig:pr} do not consider classification errors, while the results on the right side of Figure~\ref{fig:pr} require correct classification.

Under all conditions, E2ETag has higher precision than AprilTag or ChromaTag.  
E2ETag also has a higher recall at each viewing angle and for very small tags. 
AprilTag has higher recall for midrange tag sizes, however, those appear to come at the expense of precision.  
%Additionally, E2ETag is much more skewed towards succeeding when the marker is large. One advantage of E2ETag is its ability to handle varying viewing angles, as we have a better average recall and precision for each viewing angle.
%ChromaTag has significant difficulty decoding properly using their open-source marker designs and detector. Its likely that their example markers have color thresholds that do not accurately reflect the markers they introduced. Varying these color mappings and analyzing the new markers would likely be necessary to fully evaluate the performance potential of their method.
%\subsection{Augmentation Classification Accuracy}

While bit encoding methods like AprilTag and ChromaTag are designed to be class-agnostic, E2ETag only encourages this through randomized training. Therefore, to demonstrate the performance of different markers, a confusion matrix (Figure~\ref{fig:confmat}) was generated for all 30 classes.  
This test is done with Imagenet and COCO background images to adequately gather a large sample size of every marker class.  The test is designed to demonstrate performance for difficult predictions.  
%Augmentations are distributed evenly with the same parameters as Table~\ref{tab:pr_data}. 
The maximum marker size for this experiment is 32 pixels to encourage a high rate of misclassification.
The accuracy ranges from 26.78\% to 40.67\% across all the markers and the misclassification rate between any two markers ranges from 0.47\% to 7.19\%. 
Thus, some markers are more easily classified in highly challenging environments. 
However, it should be noted that misclassification is rare in practice and often, when detected, the marker is also classified correctly, as illustrated in Figure~\ref{fig:recall_dist} and \ref{fig:prec_dist}.
%-------------------------------------------------------------------------
%\vspace{-0.2cm}
%\subsection{Discussion}
%\vspace{-0.2cm}

The results of average precision and recall under various augmentations are shown in Table~\ref{tab:pr_data}.  This table presents results on the original images, as well as those images with isolated augmentations of each type applied to every image in the test set.  This includes varying blur, noise, contrast, and white balance.  
%These results demonstrate that E2ETag suffers much less performance degradation under augmentations than AprilTag or ChromaTag.  

The most significant detriments to AprilTag are additive noise, large blur, and heavy green white balance.  ChromaTag is sensitive to noise and low contrast.  
%E2ETag excels in every category and actually has a higher average recall in heavy blur, which is its worst performing augmentation, than any of AprilTag or ChromaTag's average recall.  
In contrast, E2ETag experiences a minimal loss of performance across all augmentations, with the exception of large motion blur.  
Interestingly, E2ETag has better recall under additive noise, dark contrast, and high contrast, than on the original images.  However, it does have worse classification performance under all augmentations.
\begin{table*}
\centering
\setlength{\abovecaptionskip}{10 pt}
\resizebox{1\columnwidth}{!}{%
\begin{tabular}{|c||c|c c c|c|c c c c|c c c|}
\hline
& \multicolumn{1}{|c|}{\bfseries Raw} & \multicolumn{3}{c}{\bfseries Blur} & \multicolumn{1}{|c|}{\bfseries Noise} & \multicolumn{4}{c}{\bfseries Contrast (B,W)} & \multicolumn{3}{|c|}{\bfseries White Balance}\\
% & Raw & Blur & & & Noise & Contrast (B,W) & & & & White Balance & & \\
\hline
 &  & $l$ = 5 & $l$ = 10 & $l$ = 15 & (-0.15,0.15) & (0.4,1.4) & (0.4,0.6) & (-0.4,0.6) & (-0.4,1.4) & R & G & B\\
\hline\hline
\textbf{April} & & & & & & & & & & & &\\
Precision       & 0.5093 & 0.5789 & 0.5303 & 0.4375 & 0.4775 & 0.6047 & 0.5437 & 0.8000 & 0.6477 & 0.6429 & 0.4655 & 0.5392 \\
Recall          & 0.3143 & 0.3143 & 0.2000 & 0.1200 & 0.3029 & 0.2971 & 0.3200 & 0.3886 & 0.3257 & 0.3086 & 0.3086 & 0.3143 \\
Precision Code  & 0.5093 & 0.5789 & 0.5303 & 0.4375 & 0.4775 & 0.6047 & 0.5437 & 0.8000 & 0.6477 & 0.6429 & 0.4655 & 0.5392 \\
Recall Code     & 0.3143 & 0.3143 & 0.2000 & 0.1200 & 0.3029 & 0.2971 & 0.3200 & 0.3886 & 0.3257 & 0.3086 & 0.3086 & 0.3143 \\
\hline
\textbf{Chroma} & & & & & & & & & & & &\\
Precision       & 0.6375 & 0.6780 & 0.6875 & 0.6944 & 0.0039 & 0.6923 & 0.0000 & 0.5556 & 0.2981 & 0.6000 & 0.3086 & 0.7576 \\
Recall          & 0.2914 & 0.2286 & 0.1886 & 0.1429 & 0.1143 & 0.2571 & 0.0000 & 0.2571 & 0.2743 & 0.2914 & 0.2857 & 0.2857 \\
Precision Code  & 0.5397 & 0.6545 & 0.6667 & 0.4211 & 0.0000 & 0.4595 & 0.0000 & 0.4462 & 0.2313 & 0.5526 & 0.2000 & 0.5000 \\
Recall Code     & 0.1943 & 0.2057 & 0.1714 & 0.0457 & 0.0000 & 0.0971 & 0.0000 & 0.1657 & 0.1943 & 0.2400 & 0.1600 & 0.0914 \\
\hline
\textbf{E2ETag} & & & & & & & & & & & &\\
Precision       & 0.9340 & 0.9394 & 0.9302 & 0.9583 & 0.9439 & 0.9412 & 0.9691 & 0.9528 & 0.9196 & 0.9340 & 0.9252 & 0.9333 \\
Recall          & 0.5657 & 0.5314 & 0.4571 & 0.3943 & 0.5771 & 0.5486 & 0.5371 & 0.5771 & 0.5886 & 0.5657 & 0.5657 & 0.5600 \\
Precision Code  & 0.9271 & 0.9294 & 0.9130 & 0.9464 & 0.9368 & 0.9259 & 0.9659 & 0.9412 & 0.9072 & 0.9263 & 0.9158 & 0.9271 \\
Recall Code     & 0.5086 & 0.4514 & 0.3600 & 0.3029 & 0.5086 & 0.4286 & 0.4857 & 0.4571 & 0.5029 & 0.5029 & 0.4971 & 0.5086 \\
\hline
\end{tabular}%
}
\caption{Results on real images with and without augmentations.  Precision Code and Recall Code require correct classification.  Each of the augmentations are applied individually to the entire set of Real Images.  Specifications are given for the augmentations, where blur is applied at random angles for each length and white balance augments each color channel using R:(1.3,0.7,0.7), G:(0.7,1.3,0.7), and B:(0.7,0.7,1.3).}
\label{tab:pr_data}
\vspace{-0.5cm}
\end{table*}

\begin{figure*}
\vspace{-0.8cm}
\setlength{\abovecaptionskip}{5 pt}
\begin{center}
 \begin{subfigure}[t]{\linewidth}
 \begin{center}
 \includegraphics[width=0.75\linewidth]{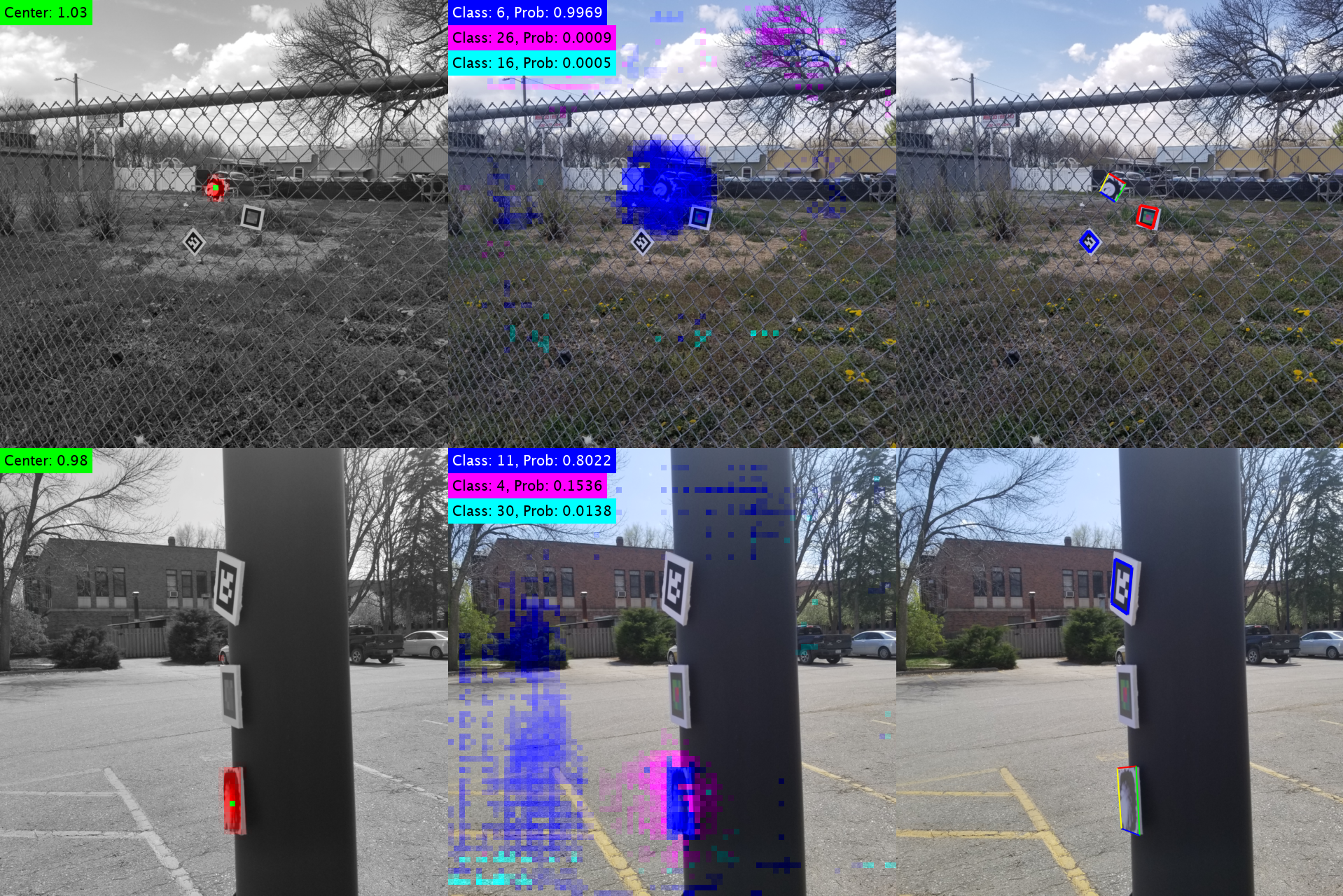}
 \caption{Easy samples with successful localization and 
 classification.}
 \label{fig:best}
 \end{center}
 \end{subfigure}
 \begin{subfigure}[t]{\linewidth}
 \begin{center}
 \includegraphics[width=0.75\linewidth]{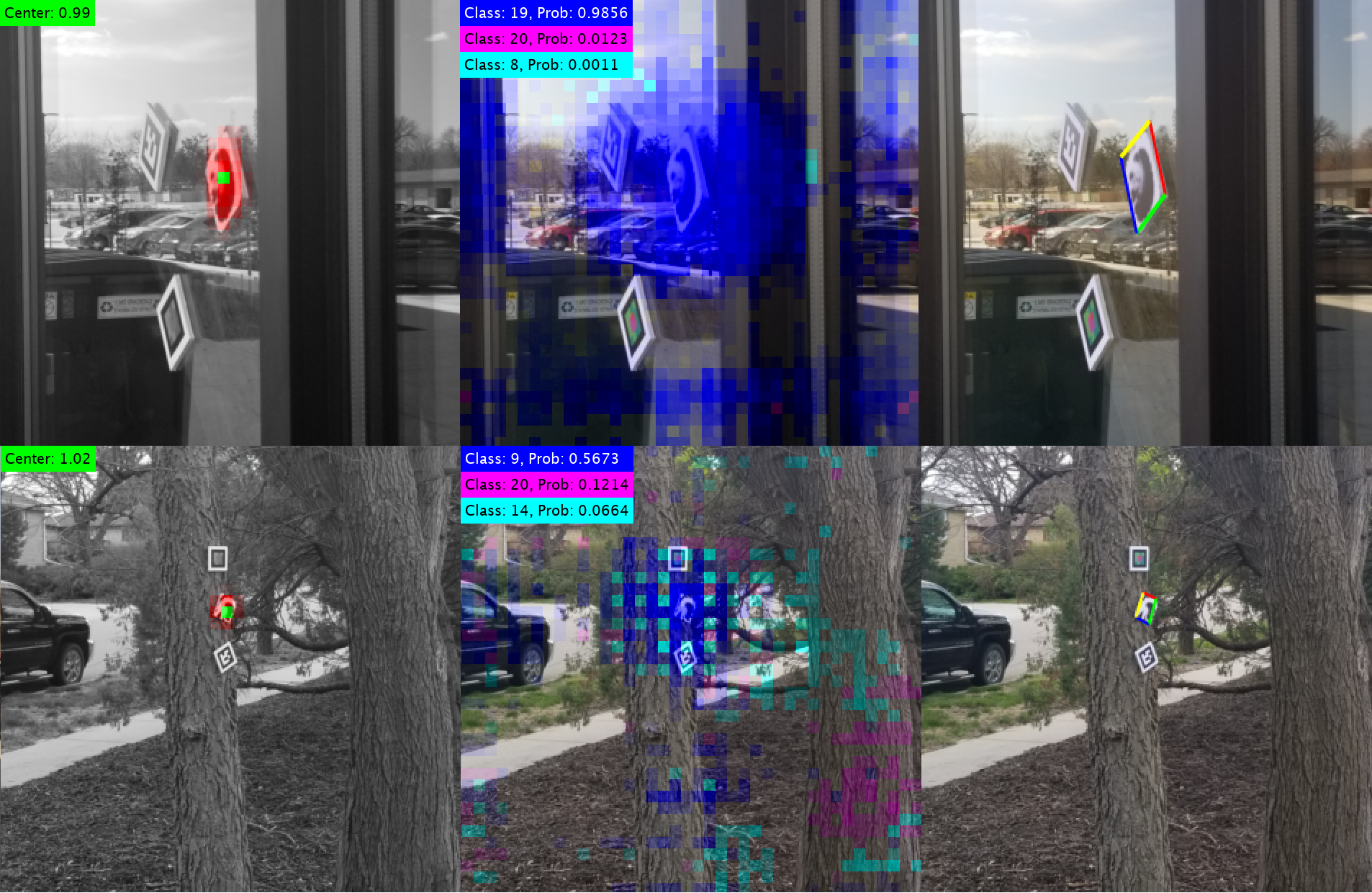}
 \caption{Difficult samples with successful localization and classification shown at 200\% crop.}
 \label{fig:mid}
 \end{center}
 \end{subfigure}
 \begin{subfigure}[t]{\linewidth}
 \includegraphics[width=1.0\linewidth]{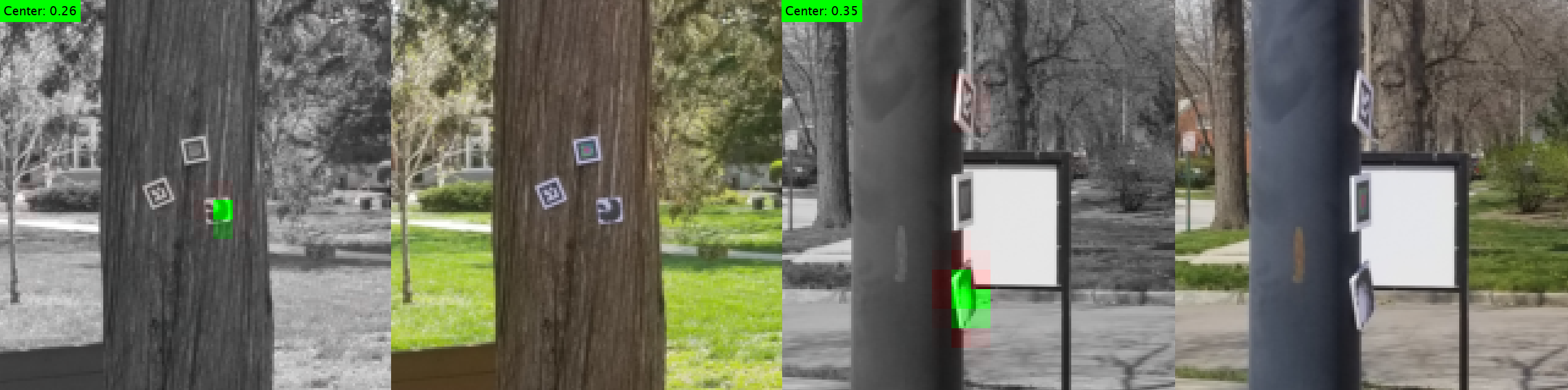}
 \caption{Samples with failed detections shown at 400\% crop.}
 \label{fig:worst}
 \end{subfigure}
\caption{Leftmost images in each row illustrate the detection channel (red with peak shown in green). The center images of each row illustrate the three largest classification channel output predictions (blue, magenta, cyan). 
The rightmost images of each row illustrate the projective transformation as colored lines around the marker and  AprilTag (blue lines) and ChromaTag (red lines) detections are shown, when detection was successful.
%This presentation is ideal and the method succeeds in all aspects. 
%In the left column of (a) designate the image with detections. The middle column contains the $80\times80$ grid containing localization values. They are overlaid in red of the leftmost grayscale image, the green grid indicates the highest value and is the grid used to predict the transform. If the center value is less than the threshold, the top 5 highest values are shown in green instead. The right column contains the $80\times80$ grid containing the correct class confidence value. The grid is overlaid in blue of the leftmost grayscale image. 
%Challenging images are shown in (b) with a 400\% crop.  The method succeeds with the left and center image, but fails (detection score $< 0.5$) to detect the right image with a tag that is approximately 12$\times$12 pixels in size.%
}
\label{fig:bestworst}
\end{center}
\vspace{-1.0cm}
\end{figure*}

To illustrate two ideal success cases, the network outputs are visualized in Figure~\ref{fig:best}. The $80\times80$ detection and classification outputs are upsampled and overlaid on the image. Two additional cases for difficult detections under challenging presentations where the method succeeds are shown Figure~\ref{fig:mid}. In both of these challenging cases, AprilTag and ChromaTag fail to detect the markers.
%The best results typically occur when the marker is captured at about 1 meter, as in Figure~\ref{fig:best}.  
%Two examples of particularly difficult cases where the method succeeds are shown in the left and center images in Figure~\ref{fig:worst}. 
%This scenario is common when the marker is 3 meters away.  
The images in Figure~\ref{fig:worst} show examples of failed detections where the marker localization does not exceed the required threshold of 0.5. These failures are likely due to the relatively small size of the tags in the image, which are contained within a 12x12 and 18x18 pixel area.
\vspace{-0.2cm}
\subsection{Hardware and Processing Time}
\vspace{-0.2cm}
The method was implemented with MATLAB using the Deep Learning Toolbox.  The computer used for training and forward inference has an Intel i9-9900K 8-core CPU, NVIDIA RTX2080ti GPU, and 16 GB of DDR4 RAM.  Detection on a 640$\times$640 frame operates at 10 frames per second.  ChromaTag and AprilTag, both explicitly designed for efficiency, are considerably faster at 900 fps, and 50 fps, respectively.
%In this case, the network made zero detections.  Since the localization prediction doesn't exceed the threshold, the network does not make a bounding box or class prediction and this exemplifies a false negative case.

\vspace{-0.2cm}
\section{Conclusion}
\vspace{-0.4cm}
By training both the marker designs and the detector together under challenging conditions, the method proposed method is able to outperform existing methods at detection and classification. 
The improvements are especially pronounced for challenging scenes and when images are corrupted by poor exposure, motion blur, and noise. 
It is noteworthy that the detector was never trained on real images, where the tag was physically placed in the scene, and it is likely that performance would improve even further if the tag designs were fixed and the detector was fine-tuned on real images.

The method is flexible and allows for a wide range of modifications. 
For example, it would be trivial to change the number of distinct markers, the shape of the markers, or even to place fixed designs into the markers.
More generally, the proposed method provides a framework for using deep neural networks to design objects that can be placed and easily detected in the real world.
%Future work on E2ETag includes new implementations into the model design of the network.  
%Implementing random noise maximum values across $0\%,10\%,25\%$ may encourage the network to generate robust markers. 
%It is likely worthwhile to investigate the effectiveness of using different detector backbones.
%This design uses a pre-trained ResNet18 to save training time, attempting to use a deeper network such as ResNet-50 could produce better results.  
%Training the detector multiple times from randomly initialized weights has the potential to improve the marker design and prediction accuracy. Also, to improve localization, adding RGB color channels to the markers might allow the marker to stand out with more certainty as the network may be capitalizing on unnatural hue gradients for detection. Color designs may also introduce more diverse patterns. 
%-------------------------------------------------------------------------
%-------------------------------------------------------------------------
\newpage
{\small
\bibliography{egbib}
}

\end{document}